\documentclass[letterpaper]{article}
\usepackage{amsmath}
\usepackage{amssymb}
\usepackage{natbib,alifeconf}  %% The order is important
\usepackage{url,hyperref,cleveref}
\usepackage{booktabs}
\usepackage{multirow}
\usepackage[dvipsnames]{xcolor}
\usepackage{stmaryrd}
\usepackage{float}

\title{Can Media Act as a Soft Regulator of Safe AI Development? %for cooperation to thrive between creators and users? 
\\ A Game Theoretical Analysis}

\author{
    Henrique Correia da Fonseca$^{1,*}$, Ant\'onio Fernandes$^{1}$, Zhao Song$^{2}$, Theodor Cimpeanu$^{3}$,
    \\{\Large Nataliya Balabanova$^{4}$, Adeela Bashir$^{2}$, Paolo Bova$^{2}$, Alessio Buscemi$^{5}$, Alessandro Di Stefano$^{2}$,}
    \\{\Large Manh Hong Duong$^{4}$, Elias Fernandez Domingos$^{6,7}$,  Ndidi Bianca Ogbo$^{2}$, Simon T. Powers$^{8}$,}
    \\{\Large Daniele Proverbio$^{9}$, Zia Ush Shamszaman$^{2}$, Fernando P. Santos$^{10}$, The Anh Han$^{2}$, Marcus Krellner$^{3}$}
    \\
    \small $^{1}$ INESC-ID and Instituto Superior T\'ecnico, Universidade de Lisboa\\
    \small $^{2}$ School of Computing, Engineering and Digital Technologies, Teesside University\\
    \small $^{3}$ Biological and Environmental Sciences, University of Stirling\\
    \small $^{4}$ School of Mathematics, University of Birmingham\\
    \small $^{5}$ Luxembourg Institute of Science and Technology\\
    \small $^{6}$ Machine Learning Group, Universit\'e libre de Bruxelles \quad
    $^{7}$ AI Lab, Vrije Universiteit Brussel\\
    \small $^{8}$ Division of Computing Science and Mathematics, University of Stirling\\
    \small $^{9}$ Department of Industrial Engineering, University of Trento\\
    \small $^{10}$ University of Amsterdam\\[2pt]
    \small $^{*}$ Corresponding author: \texttt{henrique.c.fonseca@tecnico.ulisboa.pt}
} 

\begin{document}

\maketitle

\begin{abstract}
    When developers of artificial intelligence (AI) products need to decide between profit and safety for the users, they likely choose profit.
    Untrustworthy AI technology must come packaged with tangible negative consequences. Here, we envisage those consequences as the loss of reputation caused by media coverage of their misdeeds, disseminated to the public. We explore whether media coverage has the potential to push AI creators into the production of safe products, enabling widespread adoption of AI technology. We created artificial populations of self-interested creators and users and studied them through the lens of evolutionary game theory. Our results reveal that media is indeed able to foster cooperation between creators and users, but not always. Cooperation does not evolve if the quality of the information provided by the media is not reliable enough, or if the costs of either accessing media or ensuring safety are too high. By shaping public perception and holding developers accountable, media emerges as a powerful soft regulator -- guiding AI safety even in the absence of formal government oversight.
\end{abstract}

% If sharing code / data, anonymize your repository and paste the link here.
% Example of anonymizing sevice for github: https://anonymous.4open.science/
% delete this line if not needed
Data/Code available at:\\\url{https://anonymous.4open.science/r/media-AI-governance-0752}

\section{Introduction}

In May 2024, Google introduced a feature that used Artificial Intelligence (AI) to give short answers to search prompts. Shortly thereafter, a tweet went viral \citep{McMahon2024}, showing how the prompt ``cheese not sticking to pizza" produced an answer containing the following suggestion: ``You can also add about 1/8 cup of non-toxic glue to the sauce to give it more tackiness." \citep{Kelly2024}. While this particular suggestion was so obviously ridiculous that it hopefully caused mostly laughs rather than severe medical consequences, it highlighted a substantial issue of AI technology -- can we trust it, and is it safe? 

To ensure AI safety, governments try to work towards effective regulations, such as the European Union with their ``Ethics guidelines for trustworthy AI" \citep{EU-reg}. Governmental intervention traditionally plays a role in ensuring that consumers are protected, and recently Evolutionary Game Theory (EGT) \citep{sigmund2010calculus,hofbauer1998evolutionary} models have shown how regulation of AI technology can incentivise the safe adoption of AI \citep{han2019modelling,han2020regulate, alalawi2024trust,cimpeanu2022artificial,bova2023both}. However, these models also showed some drawbacks, namely how over-regulation can prevent the adoption of valuable AI technology \citep{han2022voluntary}.

The anecdote from the beginning shows another path: a form of regulation by the media.
The story of the pizza glue was first spread via social media, then by traditional media outlets, reaching many people and making them aware of the problem. In addition, the media backlash caused Google to adjust and improve its feature.
Media has played a role in the safety of other products as well. For example, the Guardian reported that Apple contractors had been listening to confidential Siri recordings, some of which contained highly sensitive private information~\citep{Hern2019}. The resulting public backlash pressured Apple to apologise and adopt an opt-in system for reviewing recordings. Media has even influenced the inner workings of companies. For example, a Vox investigation uncovered contractual clauses that prevented departing OpenAI employees from making disparaging remarks about the company’s practices; following the media scrutiny, these clauses were removed~\citep{Piper2024}.
Concomitantly, surveys have shown that the media plays an important role in the perception of consumer risks \citep{Cao2020,Zhang2022}, which is also true for AI \citep{yang2023ai}. 
On the other hand, some research has also highlighted that an overly optimistic media coverage can be linked with increased adoption of unsafe products \citep{Melero-Bolanos2025}. The influence of the media might not always lead to the desired effects.

Artificial life researchers have long been fascinated by how complex systems self-organise, adapt, and maintain stability through decentralised interactions \citep{bedau2003artificial,powers2018modelling,krellner2021pleasing,sayama2015introduction,gershenson2020self}.
One of the most powerful mechanisms that enable such coordination is indirect reciprocity (IR)~\citep{Boyd1989, nowak2005evolution}, whereby agents cooperate based not on direct experience, but rather on reputational cues. 
While this principle has been studied extensively in biological systems, its implications for socio-technological systems and safe AI remain envisioned~\citep{paiva2018engineering}, but underexplored \citep{Xia2023, Jsang2007,Nowak1998,hammond2025multiagentrisksadvancedai}. 
Moreover, how the reputational information that is crucial to indirect reciprocity is transmitted and by whom remains an open challenge to the study of reputation-based cooperation~\citep{Hilbe2018, Santos2018,Sommerfeld2007}.

In this work, we leverage the lens of artificial life to investigate whether media -- acting as a decentralised, stochastic purveyor of reputational information -- can serve as a soft regulator in the development and adoption of artificial intelligence. 
Much like costly signalling systems in nature, media commentary provides noisy but influential feedback that shapes behaviour, not through coercion, but through perception. We model these dynamics as an evolving game between creators, users, and media agents, exploring how reputation-based safety might emerge in the real-world challenge of AI governance. In doing so, we bridge the study of artificial life with societal evolution in the context of technological norms of adoption. 

EGT models have already indicated that media -- also referred to as the commentariat in prior work -- can foster safe AI adoption in the context of governmental regulation \citep{balabanova2025media}. We therefore seek to study in particular whether media alone could enable similar results. By this, we mean that users choose to adopt AI technology and that AI creators choose to develop safe AI products. The media acts by flagging creators as safe or unsafe, hence enabling users to make an informed decision. However, information from the media can never be perfect, nor does it come for free. Taking all these factors into account, we created a model to predict the behaviour of users and creators in the presence of two distinct media outlets, differing in their cost and accuracy. In principle, our model could be seen as a general model of the role of investigative journalism as a soft regulator, without being specific to any one industry. However, unlike other high-risk industries, AI currently lacks widespread formal regulation. It is this case -- high-risk industry lacking effective formal regulation -- that our model directly addresses. We use this model to address the pressing question of to what extent media could provide incentives that fill the AI regulatory gap, while formal regulations are still waiting to be developed and enforced.

\section{Model and Methods}

\begin{figure}[t]
    \centering
    \includegraphics[width=1\linewidth]{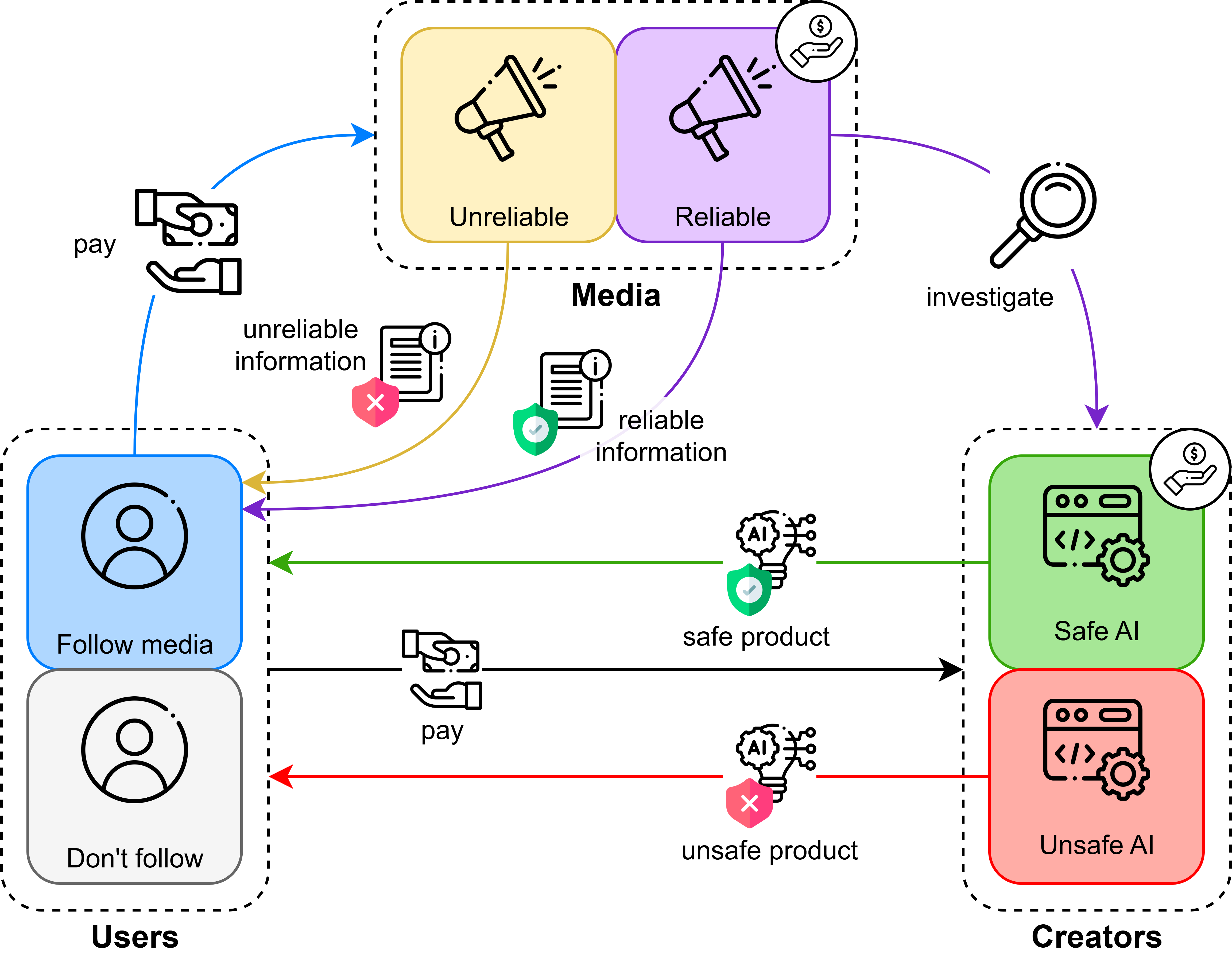}
    \caption{\textbf{Visual description of the AI regulatory ecosystem}. Users decide whether or not to use AI products, incurring a cost for adoption. For this decision process, they can choose to follow media recommendations by paying a small amount in exchange for their information about creators’ strategies. Meanwhile, creators decide whether to create safe or unsafe technology; safe technology further involves additional costs. Media can monitor the creators' decision-making process, provided that commentators invest in obtaining more reliable information.}
    \label{fig:model_overview}
\end{figure}

We use a two-population model consisting of $N_C$ creators and $N_U$ users. Users and creators evolve by updating their strategies over time. Additionally, we consider two different types of media commentators -- reliable and unreliable. Media acts as surveillance agents that monitor creators' strategies regarding AI safety practices and relay this information to users. They do not represent an evolving third population, but rather represent entities that users can choose between to get information from.

The game is played in consecutive rounds where one user meets one creator. 
The creator can choose to defect by providing an unsafe AI product or to cooperate by creating a safe one. Their cooperation is better for the user, but entails additional costs for the creator ($c_c>0$). This cost can represent increased development or production demands, and also increased effort to meet regulatory or voluntary safety requirements. %(such as disabling certain features) 
Users decide whether to cooperate by adopting this AI product or to defect by refusing to do so. Adopting always grants a benefit to the creator ($b_c>0$), but users only gain from adopting a safe product ($b_u>0$) and instead lose by adopting an unsafe one ($c_u>b_u$) (see Table~\ref{tab:parameters} for reference of all parameters).

\begin{figure}[h]
    \centering
    \includegraphics[width=1\linewidth]{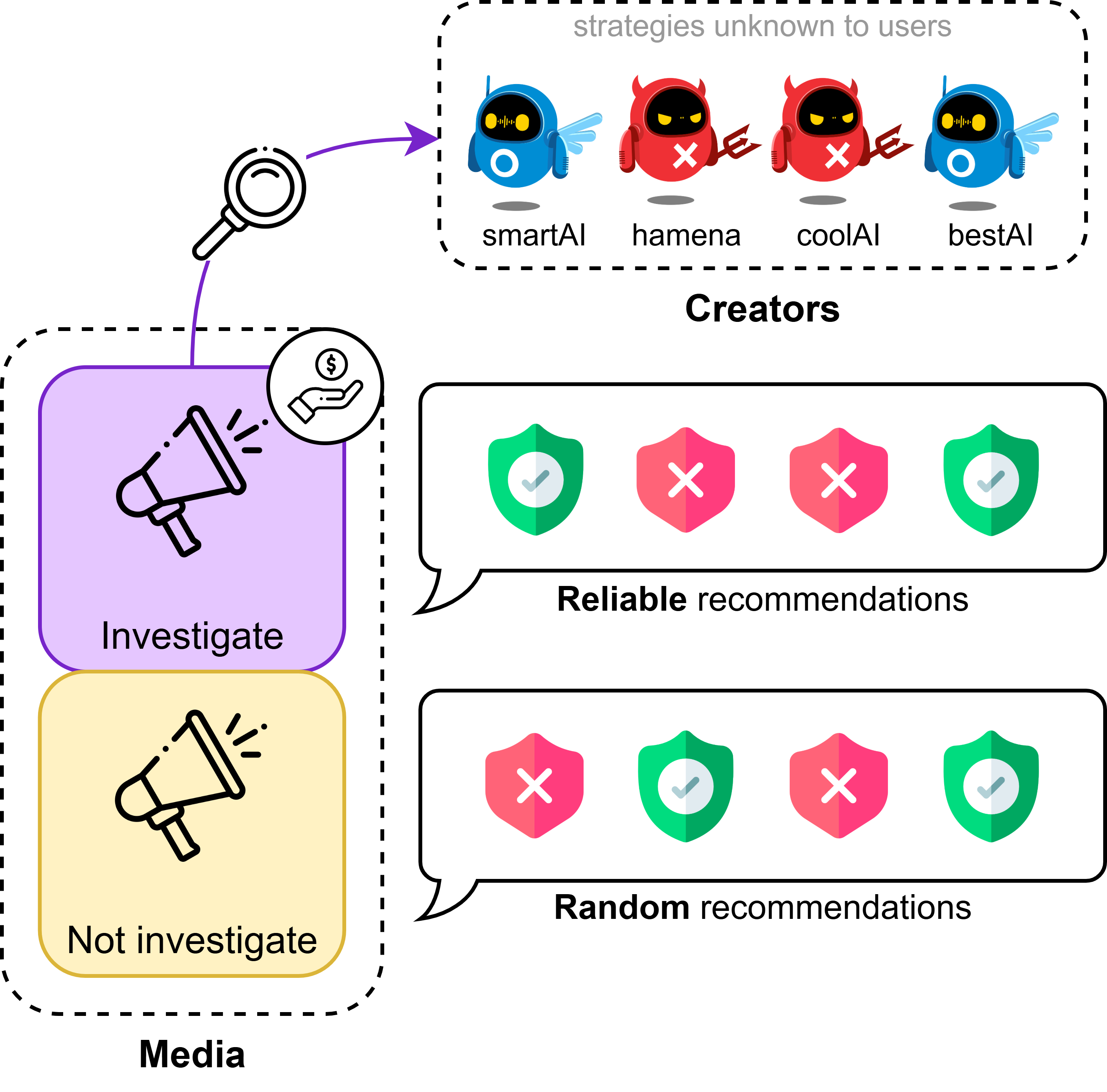}
    \caption{\textbf{Media recommendations}. Media that investigate creators' strategies can provide more reliable recommendations, although incurring an extra cost. Media that do not investigate will provide random recommendations to users.}
    \label{fig:commentariat_recommendations}
\end{figure}

Both users and creators can apply unconditional strategies: always cooperate and always defect. For the creators, we refer to these as $C$ and $D$, whereas for the users as \textit{AllC} and \textit{AllD}. For the latter, we introduce additional strategies that are based on the recommendation of a media source. We implement two types of media that they can follow, a good one (\textit{GMedia}) and a bad one (\textit{BMedia}). The good media performs thorough investigations and consequently has a chance $q$ to identify the strategy of a creator correctly (and $1-q$ to identify them incorrectly). Relying on the good media will cost the user ($c_i>0$). This represents the cost the media incurs for investigating the AI creator, which it passes on to the user in one way or another. Bad media does not perform thorough investigations of creators and so comes with no costs (see Figure \ref{fig:model_overview}). But its chance is set to $q=0.5$, meaning that it gives random recommendations (see Figure \ref{fig:commentariat_recommendations}).

\begin{table}[!h]
    \centering
    \caption{Description of the key parameters of the model.}
    \begin{tabular}{c|p{6.5cm}}
        \toprule
        \textbf{Parameter}  & \textbf{Explanation} \\
        \midrule
        $b_{u}$ & benefit a user receives when adopting a safe technology \\
        $c_{u}$ & cost a user incurs when adopting an unsafe technology \\
        $b_{c}$ & benefit a creator receives when their technology is adopted \\
        $c_{c}$ & additional cost of creating safe AI (the cost of creating unsafe AI is normalised to 0) \\
        $c_{i}$ & cost of  informed recommendation (for user or commentator)\\
        % $c_{P}$ & cost of producing safe AI technology (for the creator)\\
        $q$ & probability that the recommendation of a commentator is \emph{correct}\\
        \bottomrule
    \end{tabular}
    \label{tab:parameters}
\end{table}

The resulting payoffs are shown in Table \ref{tab:payoff_matrix}. Because the media gives probabilistic recommendations, the user behaviour is probabilistic as well, and the values are averaged for each possible interaction. %Note that because we consider interactions between two different populations (creators and users), players obtain different rewards from the same interaction and the payoff matrix is asymmetrical.  
These payoffs will be used for the replicator dynamics (see below) to determine how the strategies evolve in infinitely large populations. To supplement and verify these results, we also run agent-based simulations of finite populations. %Therein, we model the actual decision of the media and the users, but their average behaviour will remain the same.  

%\henrique{As the proposed model considers interactions between members of two different populations (creators and users), players obtain different rewards from the same interaction.
%Thus, we obtain an asymmetrical payoff matrix, as given in Table~\ref{tab:payoff_matrix}.
%Since the role of media in this model is to add another probabilistic layer to these payoffs, we model rewards of commentariat-heeding users as average payoffs: a \textit{BMedia} user will cooperate with safe creator in average 50\% of the time, whereas the probability of a $GMedia$ user cooperating with a safe creator is that of $q$.
%Thus their average expected payoffs are $0.5*b_u$ and $q*b_u-c_i$, respectively (accounting for an added cost of investigation).
%The same rationale applies to media-heeding users when facing an unsafe creator.
%}

\begin{table}[h]
    \centering
    \caption{\textbf{Payoff matrix of our game theoretical model involving users and creators}. Columns $\Pi^U$ and $\Pi^C$ represent the payoffs for users and creators, respectively, for a given combination of strategies.
    The \textit{Unsafe} and \textit{Safe} terms represent the defect and cooperate strategies from creators, respectively.
    \textit{AllD} and \textit{AllC} represent user strategies that always and never cooperate with creators, respectively, while \textit{BMedia} and \textit{GMedia} stand for deferring the decision to the recommendation of a low-quality and a higher-quality media outlet, respectively.}
    \begin{tabular}{c|c|c|c}
        \toprule
        Creator & User & $\Pi^U$ & $\Pi^C$ \\
        \midrule
        \multirow{ 4}{*}{\textit{Unsafe}} & \textit{AllD} & $0$ & $0$ \\
         & \textit{AllC} & $-c_u$ & $b_c$\\
         & \textit{BMedia} & $-0.5 c_u$ & $0.5\,b_c$\\
         & \textit{GMedia} & $-(1-q)\, c_u-c_i$ & $(1-q)\,b_c$\\
         \midrule
        \multirow{ 4}{*}{\textit{Safe}} & \textit{AllD} & $0$ & $-c_c$ \\
         & \textit{AllC} & $b_u$ & $b_c-c_c$\\
        & \textit{BMedia} & $0.5\, b_u$ & $0.5\,b_c-c_c$\\
         & \textit{GMedia} & $q \,b_u-c_i$ & $q\,b_c-c_c$\\
        \bottomrule
    \end{tabular}
    \label{tab:payoff_matrix}
\end{table}

\subsection{Replicator Dynamics}
%\subsection{Replicator Dynamics}

We first introduce the so-called replicator dynamics. %In this paper, we first assume that the user population and creator population evolve according to the replicator dynamics. The replicator dynamics 
This is a widely used model in evolutionary game theory to express how the frequencies of strategies in a population that evolves over time \citep{hofbauer1998evolutionary}. 
It is based on the idea that the proportion of agents of a given strategy increases when the strategy achieves payoffs higher than the average payoff of the population $\bar{\pi}$, and decreases when achieving expected payoffs lower than that average payoff. 
We use $x_1$, $x_2$, $x_3$, and $x_4$ to denote the frequencies of AllD, BMedia, GMedia, and AllC ($x_4=1-x_1-x_2-x_3$); and $y$ and $1-y$ to denote the frequencies of $C$ and $D$.
Formally, the replicator dynamics is given by a system of differential equations
\begin{equation}
    \begin{split}
        &\dot{x}_1=x_1(1-x_1)\left[\pi_{AllD}-\bar{\pi}_{user}\right],\\
        &\dot{x}_2=x_2(1-x_2)\left[\pi_{BMedia}-\bar{\pi}_{user}\right],\\
        &\dot{x}_3=x_3(1-x_3)\left[\pi_{GMedia}-\bar{\pi}_{user} \right],\\
        &\dot{y}=y(1-y)\left[\pi_{C}-\bar{\pi}_{creator}\right],
    \end{split}
    \label{eq: ReplicatorEquations}
\end{equation}
where the average payoff of the user population is $\bar{\pi}_{user}=x_1\pi_{AllD}+x_2\pi_{BMedia}+x_3\pi_{GMedia}+x_4\pi_{AllC}$, and of the creator population is  $\bar{\pi}_{creator}=y\pi_{C}+(1-y)\pi_{D}$. Additionally, based on the Table \ref{tab:payoff_matrix}, the expected payoffs for each strategy are:
\begin{equation}
    \begin{split}
        &\pi_{AllD}=0, \\
        &\pi_{BMedia}=0.5b_uy-0.5c_u(1-y), \\
        &\pi_{GMedia}=(qb_u-c_I)y-((1-q)c_u +c_I)(1-y), \\
        &\pi_{AllC}=b_uy-c_u(1-y),\\
        &\pi_{C}=(-c_c)x_1+(0.5b_c-c_c)x_2+(qb_c\\&\hspace{10mm}-c_c)x_3+(b_c-c_c)x_4,\\
        &\pi_{D}=0.5b_cx_2+(1-q)b_cx_3+b_cx_4.
    \end{split}
\end{equation}
With these equations, we can study how the population changes over time. 
By averaging the frequencies of the strategies over time, we can determine how frequent they are, and consequently, how much cooperation is shown by the users and the creators. 

Setting the right-hand side of Equation (\ref{eq: ReplicatorEquations}) to 0 yields the potential equilibrium states, which are points where the population composition remains static.
Any homogeneous population composition (e.g. all users are $AllD$ and all creators are $D$) might be stable.
In detail, we analyse these equilibria using Lyapunov's indirect method to assess the local stability, by examining the eigenvalues of the Jacobian matrix evaluated at these points ~\citep{khalil2002nonlinear}. For non-hyperbolic cases, we utilise centre manifold theory to reduce the system's dimensionality and facilitate stability analysis~\citep{carr2012applications}.
 %If a system starts at an equilibrium state, it will remain there forever. 
In addition, we are also interested in the evolution when it diverges from equilibria. To see this, we analyse the stability of each equilibrium.  
% \footnote{https://github.com/yt-songz/ALife2025} 
%and visualize the main results in Figure \ref{fig:replicator}. 

\subsection{Agent-Based Simulations}

Next, we introduce the methods for finite populations that rely on agent-based simulations. 
Such methods are well-known in the literature, originating from statistical physics (Monte Carlo simulations)~\citep{perc2017statistical}.
We consider two well-mixed populations: users and creators, with population sizes $N_U$ and $N_C$, respectively.
In the beginning, each creator and each user is assigned a random strategy from the set of $\{D, C\}$ and $\{AllD, BMedia, GMedia, AllC\}$, respectively. They then undergo several evolutionary time steps in which they might change their strategies as described below. 
%Two commentariat types are considered: one dubbed as ``bad" and one as ``good".
%The former provides information of quality $q_b=0.5$ (providing an accurate report of a creator's strategy with probability of 50\%) with no costs, while the latter provides information of quality $0.5 < q_g \leq 1$, with a user cost of $c_i$.
%Thus, when facing a creator with strategy $C$ (producer of safe AI), a user following some kind of media cooperates with probability given by the $q$ value of that media source.
%Users engage with creators in games of cooperation ruled by the Payoff Matrix in Table~\ref{tab:payoff_matrix}, where they accumulate fitness under selective pressure.

%\subsubsection{Evolutionary Dynamics:}

In each evolutionary step, a user and a creator are randomly selected to update their strategy (one after the other). We will describe the process for a generic player (user or creator), since the process is essentially the same.
With probability $\mu$, the player updates their strategy through mutation, randomly exploring a novel (different) strategy. 
Note that users and creators may have different mutation probabilities $\mu_u$ and $\mu_c$.
With complementary probability $1-\mu$, the focal player performs a Monte Carlo evolutionary step, whereby a second player is randomly selected (from the same population) for comparison of their respective payoffs. 
To determine these payoffs, both players engage in a number of games with randomly selected agents from the antipodal population equal to that population's size, where they accumulate payoffs, as detailed in Table~\ref{tab:payoff_matrix}.

%Applying to users: two randomly selected users play $Z_c$ games with randomly selected creators in the same current population state

%\subsubsection{Social Learning through imitation:}

After accumulating payoffs, the originally randomly selected player can update their strategy through stochastic pairwise comparison of their current fitness, as is typically done in EGT models of finite populations~\citep{traulsen2006}.
Player \textit{i} thus adopts the strategy of the second player \textit{j} with a probability given by the Fermi function
\begin{equation}
    \label{eq:fermi}
    p_{i\rightarrow j} = (1+e^{-\beta(\bar{\pi}_j-\bar{\pi}_i)})^{-1},
\end{equation} where $\bar{\pi}_j$ and $\bar{\pi}_i$ represent the average payoff of players $j$, $i$. 
The selection strength $\beta$ ranges from 0, which would mean random imitation, to infinity, where Equation~\ref{eq:fermi} essentially becomes purely deterministic.
%unit step function centered on a fitness difference between players of 0 ($f_j=f_i$).
Throughout this work, we use $\beta=1$ to ensure a strong selection strength, as is typically done in literature~\citep{Santos2021, okada2020review}, except where explicitly stated otherwise. For average results, we repeat each simulation $R=100$ times. 
A generation corresponds to $N_U+N_C$ discrete time steps, where a strategy update may occur, allowing for, on average, all members of both populations to evolve.
Each simulation runs for $G$ generations.

%\subsubsection{Average Cooperation Rate ($\eta$):}

%The average cooperation ratio in run \textit{i} ($\eta_i$) is computed by dividing the total number of cooperative acts from users $C^U_i$ and creators $C^R_i$ by the total number of games ($K_i$) twice (since agents engage in a 2-player simultaneous game), such that:
%\begin{equation}
%    \eta = R^{-1} \sum^R_{i=1} \eta_i = R^{-1} \sum^R_{i=1}\frac{C^R_i+C^U_i}{2K_i}.
%\end{equation}
%This metric is only recorded after allowing for an initial convergence period of 10\% of the total number of generations.
%We also independently record $\eta^U_i=R^{-1}\sum^R_{i=1}C^U_i/K_i$ and $\eta^C_i=R^{-1}\sum^R_{i=1}C^C_i/K_i$ as the average cooperation ratios of users and creators, respectively.

\section{Results}

The goal of our investigation is to test whether media can cause both users and creators to cooperate, resulting in the adoption of safe AI technology. We try to answer this question in different ways. We first study the behaviour of the evolving populations over time. For this, we use the replicator dynamics to run numerical simulations, which we subsequently compare with simulations of finite populations using our agent-based model. We also use the replicator dynamics to analytically study system equilibria in infinite populations. 

To study the evolution of strategies over time, we initialise the population with an equal distribution of all strategies ($25\%$ for each user strategy and $50\%$ for each creator strategy). 
We study the evolution of these populations in different settings, comprehensively exploring the parameter space. 
Specifically, we focused on varying: \textbf{(i)} the quality of media predictions ($q$); \textbf{(ii)} the cost of consuming good media for the users ($c_i$); and \textbf{(iii)} the surplus costs which must be covered by safe AI creators while remaining competitive ($c_c$).

%We were especially interested in the impacts of varying  $q$ (to see how good the information of the media must be), $c_i$ (to see how costly good media can be for the user) and $c_c$ (to see how much extra costs the safe AI creators can bear and still remain competitive). 

%We show replicator dynamics are shown in Figure \ref{fig:resrep}. 

We are especially interested in one metric -- the average cooperation rate $\eta$, as it captures cooperation in both populations. This can be computed using the frequencies of the strategies and their average behaviour (compare to Table \ref{tab:payoff_matrix}). For example, a BMedia user ($x_2$) cooperates if they receive information that the creator is trustworthy, which is equivalent to a coin flip. A GMedia user, on the other hand, cooperates in one of two cases: if they meet a cooperating creator ($y$) and get a true signal ($q$), or when they encounter a defecting creator ($1-q$) and receive a wrong signal ($1-q$). Unconditional cooperators, be it users ($x_4$) or creators ($y$), naturally always cooperate. The average cooperation rate is thus
\begin{equation}
    \eta = \frac{ y + 0.5x_2+(qy+(1-q)(1-y))x_3+x_4 }{2}.
\end{equation}

\begin{figure*}[th!]
\centering
    \begin{minipage}[b]{0.33\linewidth}
        \centering
        \includegraphics[width=\linewidth]{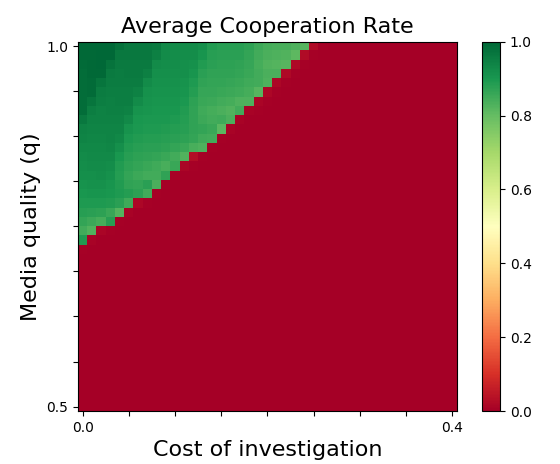}
    \end{minipage}
    \hfill
    \begin{minipage}[b]{0.33\linewidth}
        \centering
    \includegraphics[width=\linewidth]{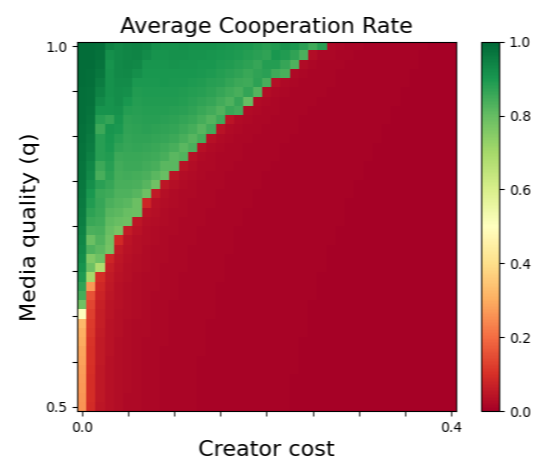}
    \end{minipage}
    \hfill
    \begin{minipage}[b]{0.33\linewidth}
        \centering
    \includegraphics[width=\linewidth]{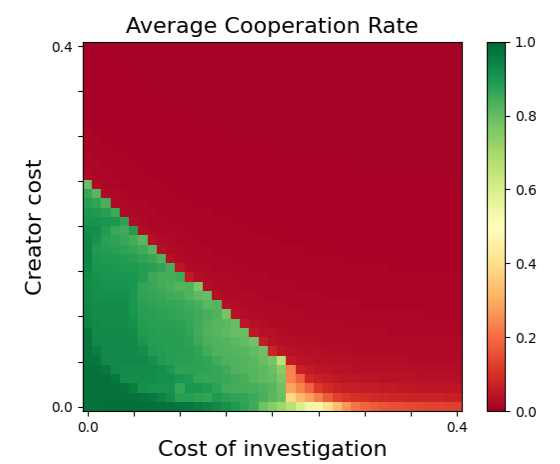}
    \end{minipage}
    \hfill
    \caption{\textbf{Average cooperation ratio $\eta$ via replicator dynamics}, across parameters of interest.
    If not varied, $q = 0.9$, $c_i = 0.1$, and $c_c=0.1$, other parameters: $b_c = 0.4$, $b_u = 0.4$, $c_u = 0.8$.}
    \label{fig:resrep}
\end{figure*}

\begin{figure*}
    \centering
    \begin{minipage}[b]{0.31\linewidth}
        \centering
        \includegraphics[width=\linewidth]{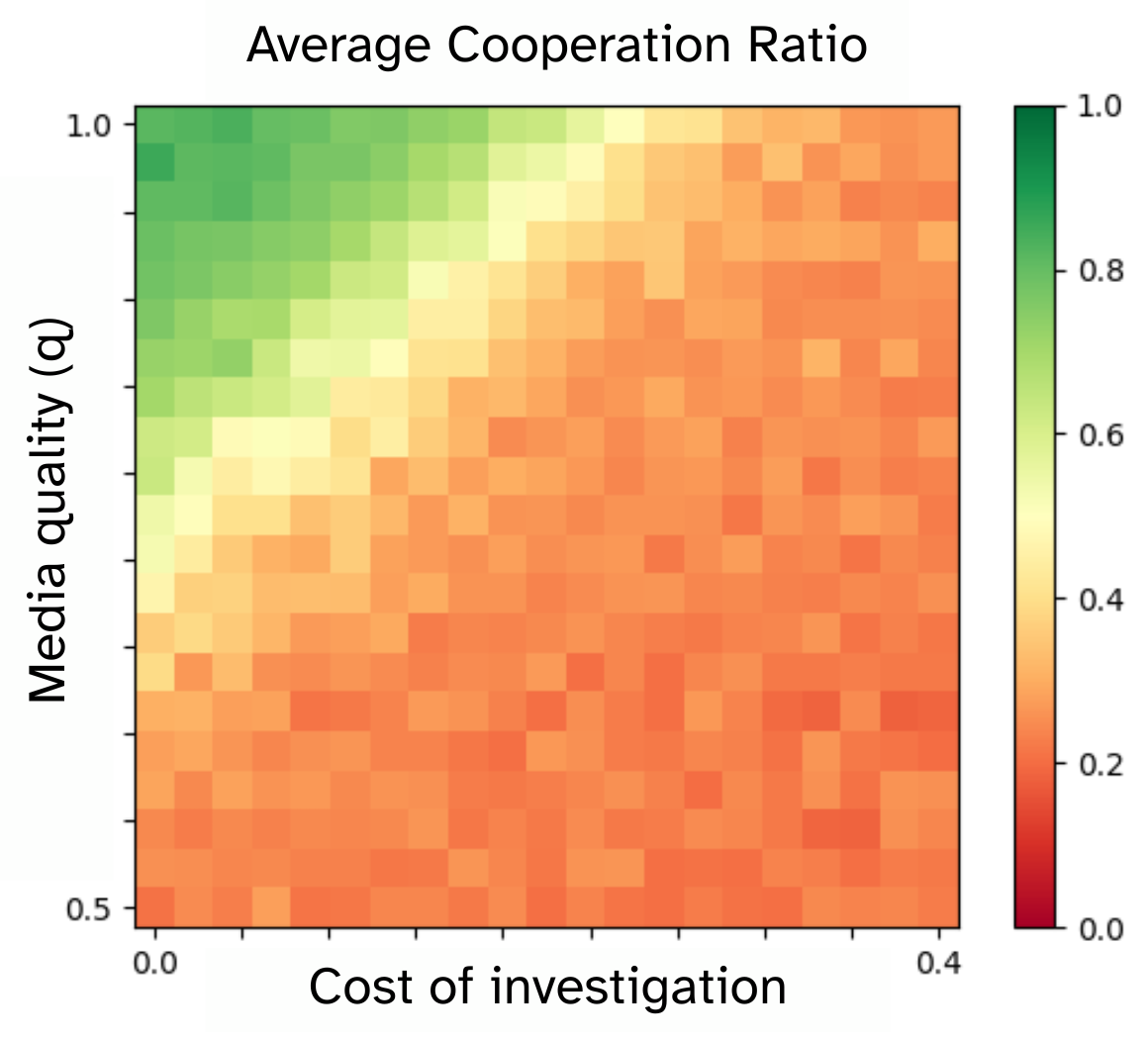}
    \end{minipage}
    \hfill
    \begin{minipage}[b]{0.31\linewidth}
        \centering
    \includegraphics[width=\linewidth]{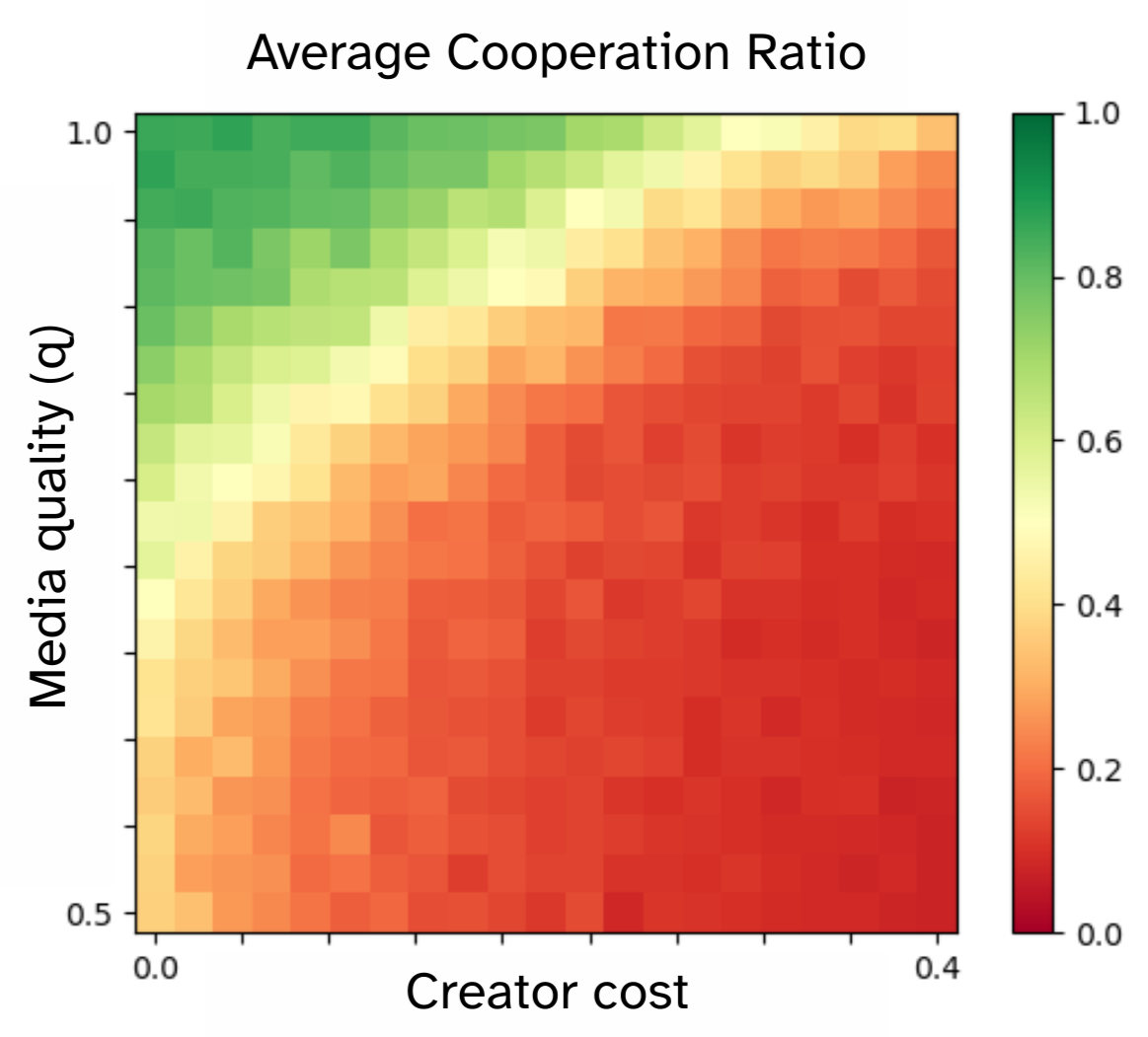}
    \end{minipage}
    \hfill
    \begin{minipage}[b]{0.31\linewidth}
        \centering
    \includegraphics[width=\linewidth]{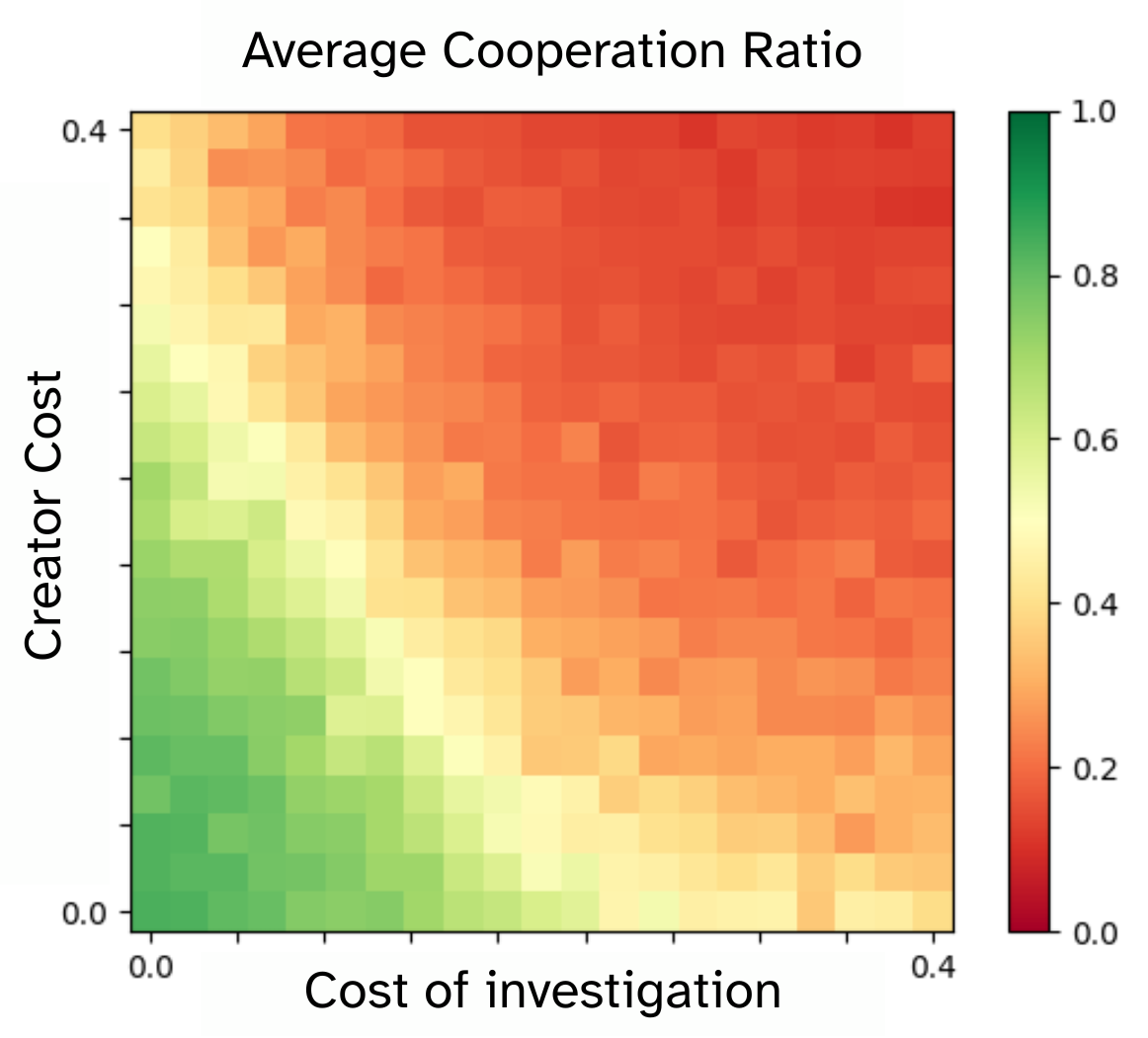}
    \end{minipage}
    \hfill
    \caption{\textbf{Average cooperation ratio $\eta$ via agent-based simulations,} across parameters of interest. Each data point shows the average cooperation ratio $\eta$ averaged over $R=100$ runs. All parameters are set to the same ones as Figure~\ref{fig:resrep}, with the additional evolutionary parameters of $G=500, N_U=100, N_C=50,\beta_C=\beta_U=1, \mu_u=1/N_U, \mu_c=1/N_C$.
    The metric $\eta$ is only computed after allowing for a converging period of $G/10$ generations.}
    \label{fig:simulation_heatmaps}
\end{figure*}

The result of the numerical simulations is shown in Figure \ref{fig:resrep}. They show the relationship between cooperation and the parameters of interest. Crucially, we observe that all three parameters have the capacity to completely collapse cooperation but also to achieve high levels of cooperation. 

Taken in pairwise isolation, we show thresholds beyond which creators always resort to unsafe AI development. Firstly, media must always tread a fine balance: not only must it maintain sufficient quality $q$ of its reports on creator behaviour, but this threshold gets stricter as the cost of investigating creators increases (Figure \ref{fig:resrep} first panel). Similarly, creator costs react in much the same fashion, imposing a certain amount of strictness on the existing commentariat (Figure \ref{fig:resrep} second panel). In other words, when either of the costs is prohibitive (either to creators $c_c$, or to media $c_i$), then media providers must provide accurate information to compensate, else the advantage they gain from user trust is overcome by lazy media. Secondly, we show a more intuitive relationship between the two costs ($c_c$ and $c_i$), as either of them can lead to a breakdown of cooperation if excessive. There is a slightly more forgiving nature to costly investigation ($c_i\lesssim0.35$) as opposed to the cost of safety ($c_c\lesssim0.25$), but cooperation can be achieved if neither threshold is overstepped. 
% \begin{figure}[]
%     \centering
%     \includegraphics[width=\linewidth]{images/fig-internal.pdf}
%     \caption{Frequency of user (top) and creator (bottom) strategies over simulation time, obtained through replicator dynamics. 
%     Persistent interdependent oscillation dynamics can be observed in user and creator populations. 
%     Shown are the frequencies of user and creator strategies over time. Parameters are set as $b_c = 0.4$, $c_c = 0.2$, $b_u = 0.4$, $c_u = 0.8$, $c_i = 0.05$, and $q = 0.9$.}
%     \label{fig:replicator}
% \end{figure}

Following this approach, we tried to replicate these findings with agent-based simulation of finite populations, where strategy evolution occurs through mutation and social learning. 
Figure~\ref{fig:simulation_heatmaps} shows similar insights to the ones obtained through replicator dynamics (Figure~\ref{fig:resrep}). 
Due to the stochastic nature of the finite models, transitions between areas of cooperation and defection are less sharp, but follow the same patterns very closely. 
This provides robustness to our main finding that media supports cooperation between users and creators, given realistic quality levels, costs of investigative media and costs of safe development.

%\subsection{Detailed Behaviour}

\begin{figure}[ht]
    \centering
    \includegraphics[width=\linewidth]{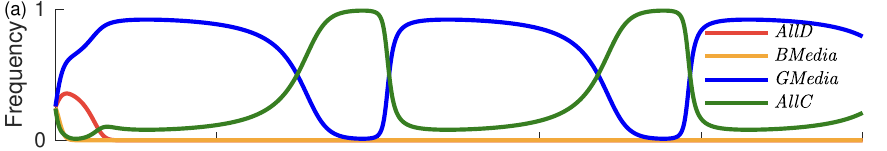}
    \vspace{15pt}
    \includegraphics[width=\linewidth]{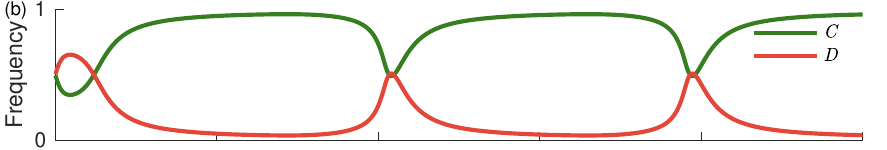}
    \includegraphics[width=\linewidth]{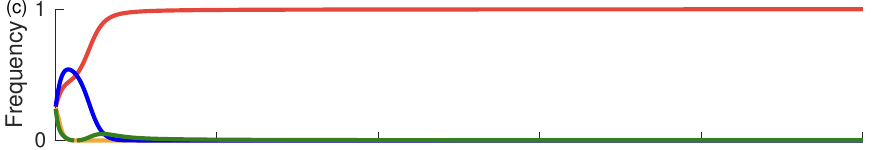}
    \includegraphics[width=\linewidth]{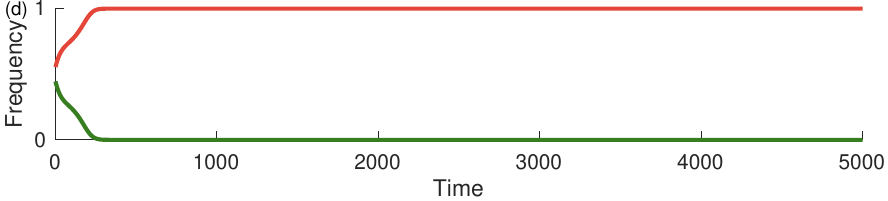}
    \caption{\textbf{Evolution of strategies over time for replicator dynamics}. 
    In the top two panels (plots \textit{a} and \textit{b}), with initially 50\% $C$ in the creator population, persistent interdependent oscillation dynamics are observed in the user and creator populations. 
    In the bottom two panels (plots \textit{c} and \textit{d}), with initially 45$\%$ $C$ in the creator population, $AllD$ dominance and $D$ dominance are observed in the user and creator populations. 
    Parameters are set as $b_c = 0.4$, $c_c = 0.2$, $b_u = 0.4$, $c_u = 0.8$, $c_i = 0.05$, and $q = 0.9$.}
    \label{fig:replicator}
\end{figure}

We then explore the stability of the infinite populations as described above. Our analysis reveals that although achieving a state of full cooperation within the user population remains an unattainable equilibrium, cooperation endures through an intriguing evolutionary dynamic characterised by the persistent oscillation between $GMedia$ and $AllC$. 

As illustrated in Figure \ref{fig:replicator}(a-b), upon the initiation of the evolutionary process, $AllD$ and $BMedia$ rapidly decline within the user population, while $GMedia$ and $AllC$ exhibit sustained oscillatory behaviour. Similarly, in the creator population, the frequencies of $C$ and $D$ also display persistent oscillations. This dynamic interplay allows $AllC$ to survive and evolve in concert with $GMedia$.

An in-depth examination of the equilibria yields a stable point in which $AllC$ dominates in the user population and $C$ dominates in the creator population. However, it is only stable if $
c_c<0, b_u>0,  c_i+b_u(1-q)>0$. We limited the parameters, such as $c_C$ to be positive, because we wanted to explore realistic scenarios. This stable point has, therefore, no relevance for our overarching question. %However, since the first inequality clearly does not hold, this equilibrium point is unstable. It's worth noting that even though full cooperation cannot manifest as a stable, steady state, its persistence as a continuous oscillation, driven by $GMedia$, underscores the crucial role of media. 

\begin{figure}[ht]
    \centering
    \includegraphics[width=\linewidth]{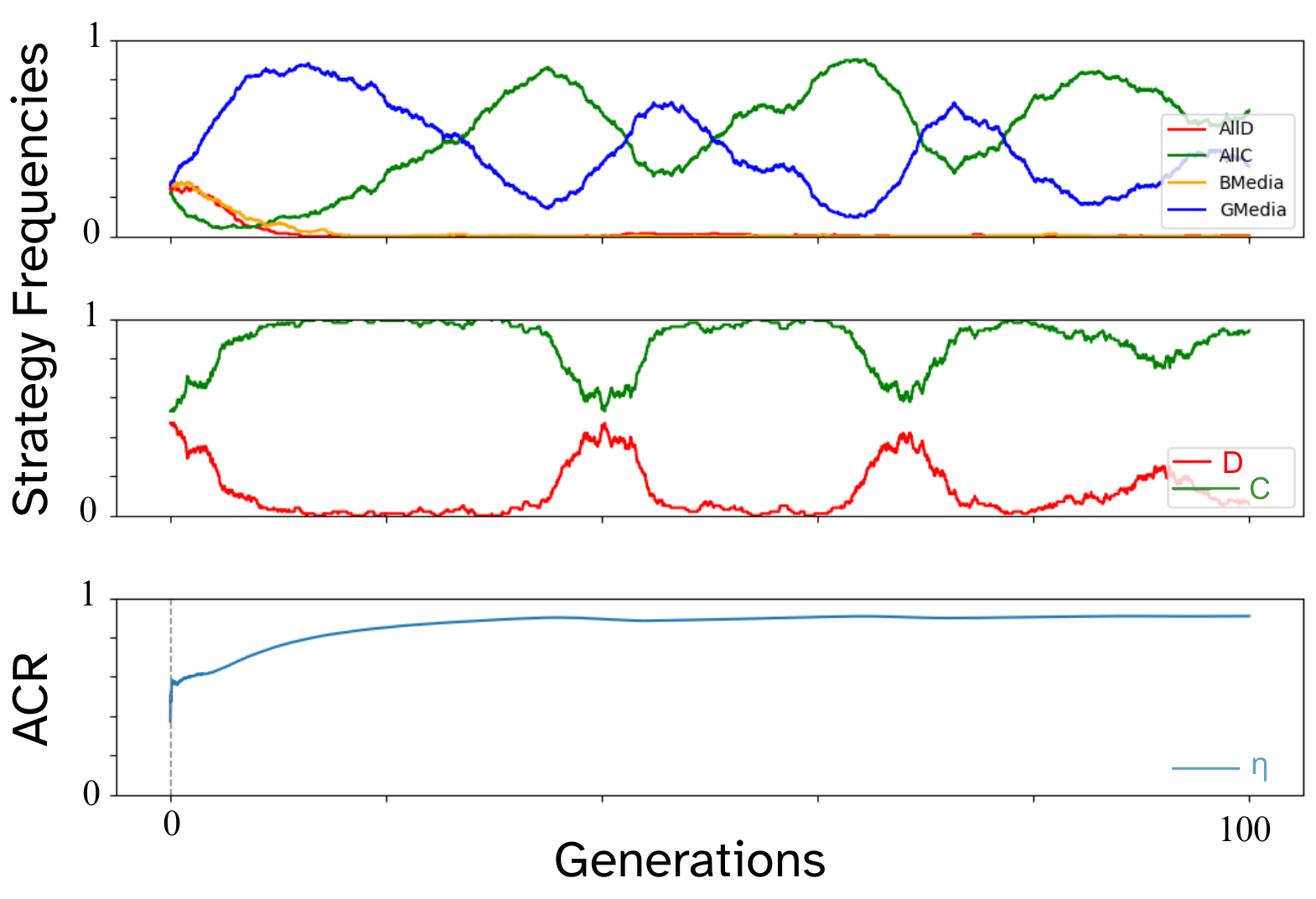}
    \caption{\textbf{Evolution of strategies and cooperation over time for the agent-based model} of user (top) and creator (middle) strategies, alongside the average cooperation ratio (ACR, or $\eta$) (bottom), for a typical run of the simulation ($R=1$) with $N_U=2N_C=100, \mu_u=1/N_U=0.005, \mu_c=1/N_C=0.01$ and other payoff values set to the same as in Figure~\ref{fig:replicator}. 
    The simulation is run for $G=100$ generations for a minute illustration of the typical evolutionary dynamics.
    User strategies $AllD$, $AllC$, $BMedia$, and $GMedia$ are pictured in red, green, yellow, and blue, respectively. Creator strategies, \textit{Unsafe} ($D$) and \textit{Safe} ($C$), are pictured in red and green, respectively.
    }
    \label{fig:simulation_cycles}
\end{figure}

However, we found the equilibrium point %$(1,0,0,0)$, 
where $AllD$ dominates in the user population and $D$ dominates in the creator population. It is stable for the conditions $c_c>0, c_u>0, c_i+c_u(1-q)>0$. They always hold for the parameter space we consider realistic. However, the existence of a stable point for defection does not mean that cooperation is impossible \citep{Imhof2005}.
%time to finish up, Marcus :) also are you ok with St Andrews affiliation?

In our model, convergence to the stable defection point seems to depend on initial conditions. Even under the same parameter settings that produce oscillatory dynamics, different starting states can drive the system to collapse into this defection-dominated equilibrium. This sensitivity highlights a possible bistable regime in the model: the same parameters can yield either persistent cooperation (via oscillations) or universal defection, contingent on the initial states of the populations, see Figure \ref{fig:replicator}(c-d). 
%Due to space limitations, detailed analyses are omitted here. 

%We have already shown that the only stable point of the population is that all creators and all users are defectors (given that all costs are non-negative). This means that starting the population in that state would yield no changes in the frequency of strategies and hence an average cooperation rate of $\eta=0$. Yet, if we start the populations in a balanced state, i.e. all strategies have the same frequency (25\%  for the four user strategies and 50\% for the creator strategies), this stable state will for many parameter settings never be reached. Instead, the population oscillates between different user and creator strategy frequencies. 

We quantify which of these two outcomes, oscillation with high cooperation or collapse to full defection, is more relevant for the answer to our research question. We therefore used the replicator dynamics to study the evolution of the population from different starting states. We use the same numerical approach as described above and vary the frequencies by steps of $2\%$, which are equivalent to 1172286 different starting states. Of these, $38.6\%$ reach the stable defection state, the others do not (parameters as in Figure \ref{fig:resrep}). Across all starting states, the measured average cooperation rate is $\eta=0.54$. We therefore consider both dynamics as relevant.

%\subsection{Agent-Based Simulations}
%maybe: For finite populations, mutations exclude truly stable states?
For our finite model, there exist no truly stable states because of mutation. However, it is interesting to compare their behaviour with the replicator dynamics. Indeed, even the stochastic agent-based model shows consistent oscillations of strategy frequency -- see Figure \ref{fig:simulation_cycles} -- for the same parameters of Figure~\ref{fig:replicator}. 
%Then, how do we obtain the levels of cooperation observed in Figures~\ref{fig:resrep} and \ref{fig:simulation_heatmaps}?
%To answer that question, we can examine the example of Figure~\ref{fig:simulation_cycles}, plotting the strategy frequencies obtained by running one simulation of the computational model with finite populations, mutations and imitation learning, for the very same payoff configuration of Figure~\ref{fig:replicator}.
%One can observe that, in the former, %figure 
%the ever-repeating phase-cycles are also present in the latter, whereby 
Cycles follow the same pattern: (i) $GMedia$ users proliferate in face of the initial population state, leading to (ii) the invasion of $AllC$ user strategies that take advantage of the highly cooperative creator population state, which in turn (iii) are easily exploited by an invading population of defective creators, (iv) counteracted by a non-zero population of $GMedia$ users whose discriminating strategy stops the rise of defective creators, leading the system back to step (ii).
Notably, users of $GMedia$ do not dominate the population. 
%This behaviour, however, hinges on behavioural exploration. Without mutations, the system quickly converges to the extinction of certain strategies, highly dependent on the initial stochastic conditions.

\begin{figure}
    \centering
    \includegraphics[width=\linewidth]{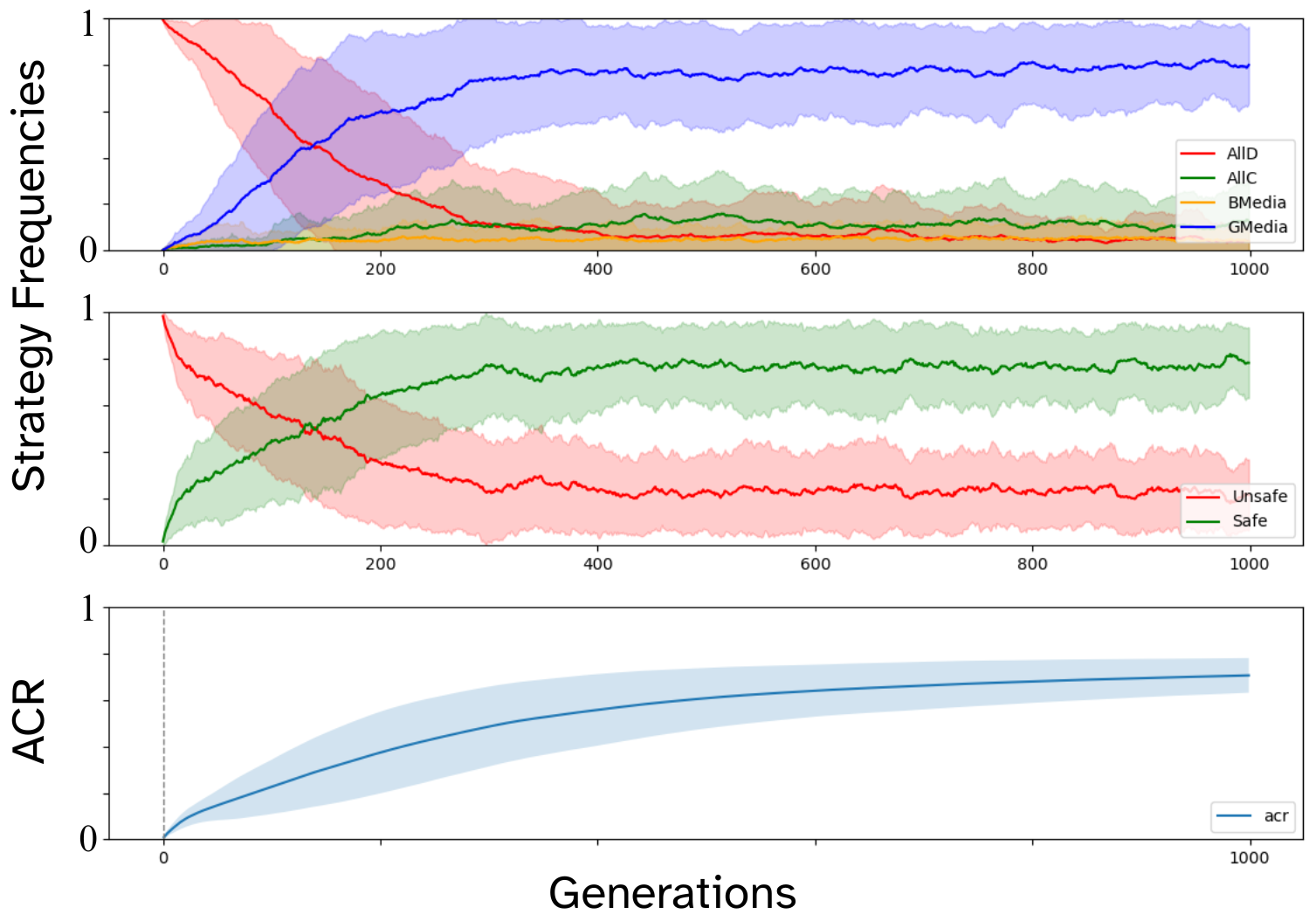}
    \caption{\textbf{Initial state with only defective strategies.} Evolution over simulation-time of user (top) and creator (strategies, as well as cooperation levels (bottom), where the initial population state is that of only AllD users and D creators.
    Colored shaded areas represent the standard deviation of averaging each metric and result.
    Results averaged over $R=100$ runs.
    Payoff parameters set to $q=0.9, b_c=0.4, b_u=0.4, c_c=0.1, c_u=0.8, c_i=0.05$ and evolutionary parameters set to $N_U=2N_C=100, \beta_C=\beta_U=1, \mu_U=1/N_U, \mu_C=1/N_C, G=1000$.
    }
    \label{fig:sim_allD_start}
\end{figure}

To see if the initial population composition has any significant effect on the simulations, we run several examples starting in the predicted stable defection state, averaging the frequencies of strategies over time (see Figure \ref{fig:sim_allD_start}). We note that the simulated populations can consistently escape this state and display high levels of cooperation. 

All the results from the computational model are robust to variations in the number of generations, population sizes, and both selection strengths and mutation rates, as long as their order of magnitude remains above a certain threshold ($\beta\geq1$ and $mu\geq0.1$).

\section{Discussion}
AI safety is a concern, because -- all things being equal -- providing safe AI products is costly, giving creators an incentive to provide unsafe AI instead \citep{han2019modelling,cohen2024regulating,hammond2025multiagentrisksadvancedai,bengio2025international}. In this paper, we show that media alone can act as a soft regulator of AI creators' development behaviour, by providing users with information about which AI products are safe to adopt. In our model, we formulated populations with two types of media -- good media, which pay investigative costs to produce a reliable signal, and bad media, which give random signals. Even if the signal by the good media is far from perfect, and even if users have to pay substantially for the service of such media, it was able to maintain cooperation (products are safely developed and users adopt the products) for a wide range of the parameter space. However, we also showed that media in some cases failed to enable cooperation. These include: when the cost of good media or of safe AI creation was too high; when the signal of the media was too noisy; or when the population started with a large majority of defection by creators and users. 
Our findings were robust to both analytical predictions and agent-based simulations. Furthermore, we found that populations with high cooperation often experience cycles of behaviour. Widespread use of good media leads to the adoption and creation of safe AI, undermining the need for good media. At this point, users blindly adopt all AI products. This, in turn, leads to an increase in unsafe AI creation that exploits the dominant naive user population, making good media valuable again, thus restarting the cycle. 

Our work is related to the large body of research on indirect reciprocity \citep{nowak2005evolution,okada2020review}. Indirect reciprocity means that if player x cooperates with player y, player x is rewarded by the cooperation of another player z, because x has built a good reputation. In such models, reciprocity can flow from any player to any other. In our model, on the other hand, reciprocity only flows from users towards creators, not vice versa nor within the populations. In standard indirect reciprocity, players are motivated to cooperate because they want to receive reciprocal cooperation. In our model, many players -- namely users -- can never receive reciprocal cooperation because their behaviour is entirely unmonitored. Instead, their cooperation is motivated by a self-interest to benefit from their own current cooperation (i.e. by benefiting from the use of AI), which they can only receive from creators that are -- coincidentally -- worth of reciprocal cooperation.  

\subsection{Limitations and Future Work}

Our model is the first of its kind, showcasing the pure effect of media on safe AI adoption. In this, we limited the concerns of safety to the immediate user of the technology. Examples that could be captured by our model include the use of Large Language Model (LLM)-based health applications by patients, which could provide the patient with inaccurate or harmful information if not correctly regulated \citep{freyer_future_2024}, or the use of AI chatbots that could leak user conversations (as reported in the media with ChatGPT's ``share'' feature; \citealp{prada_chatgpt_2025}). Our model does not, however, capture societal-wide consequences of unsafe AI development, such as plagiarism and copyright violations by LLMs, biased decisions adversely affecting minority groups \citep{wu_unveiling_2024} or widespread opinion polarisation through link-recommendation algorithms~\citep{Santos2021opinion}. 
Considering these effects would require a substantially different model and is a worthwhile extension for the future. We also envisage several other expansions to address current limitations. 

We only considered four types of users and two types of media at a time. More realistically, a whole ecosystem of different kinds of media outlets co-exists, most of which will have different levels of budgets for investigation and hence different levels of quality. Some media outlets spend millions of dollars, whereas actors on social media may only spend seconds of their time. Even more realistically, the accuracy of media is not just determined by their effort of investigation, but also by their bias. Many actors, especially in the social media realm, are either overly enthusiastic or critical about certain topics, including AI, to say nothing of the role of the political affiliation of media providers \citep{yang2023ai}.

To address this shortcoming, we propose to expand our model in the future by introducing parameters for bias and expenditure. Bias could be a determinant of the initial assumption of a media outlet about all creators (before they start their investigation). It could range from trusting every creator to distrusting them all. During their investigation, the media outlet then has a chance to discover the real value of the creator and change its opinion accordingly. This chance will depend on the individual expenditure of the media. Only if the true value is discovered will the initial assumption be overwritten. This way, a media outlet can provide valuable information but still be biased. This is especially interesting to the evolutionary dynamics, since biased media might be more profitable \citep{BARON20061}. However, increasing the set of possible media might be challenging to study.  

A second limitation is that we assumed users would only listen to a single media source. However, in real life, users can hardly avoid being bombarded by many different sources. Integrating multiple sources of information is not trivial \citep{Massaro1990}. Future research will need to define and compare heuristics for users to deal with multiple media sources that are available to them at the same time, especially if they provide contradictory information about the safety of AI. 
%form of secondary trust might play a role, a trust not in AI creators, but in AI media.

In this work, we purposely limited the possible regulation of AI safety to be done by the media, using only reputations, but no other enforcement. Realistically, such enforcement, mainly by governments, is starting to come into place and is needed to ensure safety under all circumstances. We therefore want to continue the research of artificial population with government regulation \citep{han2020regulate, alalawi2024trust,cimpeanu2022artificial,bova2023both,han2022voluntary,buscemi2025llms}, by combining it with media oversight  \citep{balabanova2025media,powers2023stuff}, using more complex artificial systems to understand the dynamics of the real struggle for AI safety. 

\section{Acknowledgments}
The authors acknowledge support by the Future of Life Institute (mini-grant for ``AI Governance Modelling Workshop organisation" by TAH), EPSRC (grant no. EP/Y00857X/1 and grant no. EP/Y008561/1), CRCRM (MR/Z505833/1), European Commission (ERC). Additionally, this project was supported by the INESC-ID (UIDB/50021/2020), as well as by the Centre for Responsible AI (CRAI) project (grant no. C645008882-00000055/510852254 and C628696807-00454142, IAPMEI/PRR). 

%\section{Appendix I}

%To illustrate the influence of the starting state, we plotted four graphs in figure []. In each, the frequency of creator strategies is varied across all possibilities, but for the users in each graph, there is only one focal strategies the frequency which is systematically varied, while the others are kept balanced (they have equal shares of the players that do not play the focal strategy). With this, we can see that cooperation is again very frequent in about 50\% of starting states. The starting frequency of GMedia has a positive impact, whereas the starting frequencies of AllD has the strongest negative impact on cooperation, and the frequency of defecting creators has a strong negative impact as well. However, if the frequency of GMedia is high, even high frequencies of defecting creators to not cause the cooperation in the population to collapse.

\footnotesize
\bibliographystyle{apalike}
\bibliography{example,bibMK,refs} % replace by the name of your .bib file

\end{document}